\documentclass{article}

\usepackage[preprint]{template}

\usepackage[utf8]{inputenc} 
\usepackage[T1]{fontenc}    
\usepackage{hyperref}       
\usepackage{url}            
\usepackage{booktabs}       
\usepackage{amsfonts}       
\usepackage{nicefrac}       
\usepackage{microtype}      
\usepackage{xcolor}         
\usepackage{multirow}
\usepackage{adjustbox}
\usepackage{amssymb}
\usepackage{amsmath}
\usepackage{physics}
\usepackage{pifont}

\graphicspath{ {images/} }

\title{Hybrid diffusion models: combining supervised and generative pretraining for label-efficient fine-tuning of segmentation models}

\author{
  Bruno ~Sauvalle \\
  Mines Paris PSL University \& EPFL\thanks{Centre for Robotics, Mines Paris. Work conducted as a visiting professor at EPFL, Computer Vision Laboratory}\\
   \And
   Mathieu Salzmann \\
   Swiss Data Science Center \& EPFL \\
  }

\begin{document}

\maketitle

\begin{abstract}

 We are considering in this paper the task of label-efficient fine-tuning of segmentation models: We assume that a large labeled dataset is available and allows to train an accurate segmentation model in one domain, and that we have to adapt this model on a related domain where only a few samples are available. We observe that this adaptation can be done using two distinct methods: The first method, supervised pretraining, is simply to take the model trained on the first domain using classical supervised learning, and fine-tune it on the second domain with the available labeled samples. The second method is to perform self-supervised pretraining on the first domain using a generic pretext task in order to get  high-quality representations which can then be used to train a model on the second domain in a label-efficient way. We propose in this paper to fuse these two approaches by introducing a new pretext task,  which is to perform simultaneously image denoising and mask prediction on the first domain. We motivate this choice by showing that in the same way that an image denoiser conditioned on the noise level can be considered  as a  generative model for the unlabeled image distribution using the theory of diffusion models, a model trained using this new pretext task can be considered as a generative model for the joint distribution of images and segmentation masks under the assumption that the mapping from images to segmentation masks is deterministic. We then empirically show on several datasets that fine-tuning a model pretrained using this approach leads to better results than fine-tuning a similar model trained using either supervised or unsupervised pretraining only.

\end{abstract}

\section{Introduction}

Image segmentation is one of the most fundamental computer vision tasks. Current state of the art segmentation models rely on supervised learning, which requires the availability of a large dataset of labeled data. However creating labeled data is far more difficult and expensive for segmentation than for image classification, since labeling a complete image requires to annotate the class of each pixel of this image. As a consequence, the development of label-efficient training methods, i.e. methods allowing to train a model using only a small number of labeled samples, has been the subject of numerous research works in the last decade. 

One can distinguish two main approaches to handling this problem:

\begin {itemize}

\item The first method is called supervised pretraining: A segmentation model is first trained  on a dataset where enough labeled data is available, and then fine-tuned on a different target dataset, where only a few labeled samples are available. 

\item The second method, called unsupervised pretraining, is to assume that a large number of unlabeled data samples of the dataset of interest is available and to pretrain a model using these unlabeled samples to learn efficient representations, by leveraging various self-supervised learning strategy, and then to fine-tune the pretrained model using the available labeled samples.

\end{itemize}

We show in this paper that it is possible to fuse these two approaches. More precisely, we take our inspiration from recent works showing that UNets pretrained as unconditional diffusion models learn representations that can be label-efficient for segmentation in the classical semi-supervised setting, when only one domain is considered. In the  setting considered in this paper, where sufficient labeled data is available in a first domain, and the model has to be adapted to a second domain where only a few samples are available,  we propose to train a UNet simultaneously as a segmentation model and a diffusion model on the first domain before fine-tuning it on the second domain where the data is scarce. Our hypothesis is that the representations learnt using this process will be more efficient for segmentation fine-tuning than if the UNet is trained only as a diffusion model or as a supervised segmentation model. 

The main contributions of this paper are the following:

\begin{itemize}
\item We introduce the concept of hybrid diffusion model and provide theoretical results showing that a trained hybrid diffusion model can be used both as a generative model for the joint distribution of the images and associated masks and as a supervised segmentation model.
\item We provide experimental results showing that fine-tuning a pretrained hybrid diffusion model is more effective than fine-tuning a similar model pretrained using supervised or unsupervised pretraining alone.
\end{itemize}

\section{Related work and background}
\label{sec:related_works}
\citet{baranchukLabelEfficientSemanticSegmentation2021} and \citet{rousseauPreTrainingDiffusionModels2023} showed that fine-tuning a pretrained class-unconditional diffusion model could give very good results for segmentation tasks when a few labeled samples are available in the classical semi-supervised setting where only one domain is considered. Both works rely on the ADM UNet architecture \citep{dhariwalDiffusionModelsBeat2021}, which combines convolutional layers and global attention layers. 

 \begin{itemize}
  \item \citet{baranchukLabelEfficientSemanticSegmentation2021}  proposed to freeze the UNet after pretraining. For fine-tuning, a separate MLP is trained in a pixel-wise manner to perform segmentation using as input the corresponding pixel feature map activations of the UNet for different time-steps, rescaled as necessary. \citet{rosnatiRobustSemisupervisedSegmentation2023a} later showed that with this method, the smallest time-steps are the most informative and proposed to share MLP weights across time-steps to improve generalization.  \citet{liDreamTeacherPretrainingImage2023a} showed that although the diffusion models  recommended for this method are large (554 M parameters), it is possible to distill these models into a smaller backbones without accuracy loss.

 \item \citet{rousseauPreTrainingDiffusionModels2023} proposed to add a linear classification layer to a pretrained time-conditioned diffusion model with a fixed condition $t=1$  and fine-tune it as a segmentation model. More recently, \citet{huangLabelefficientMultiorganSegmentation2024a} proposed a similar method, but fine-tuning only the decoder of the diffusion model and not the encoder.

\end{itemize}


These results are consistent with numerous works \citep{vincentExtractingComposingRobust2008,vincentStackedDenoisingAutoencoders, xiangDenoisingDiffusionAutoencoders2023b,youPretrainedFeaturesNoisy,mukhopadhyayDiffusionModelsBeat2023}, which have shown since 2008 that denoising is an effective pretext task for generic self-supervised learning.
 \citet{prakashLeveragingSelfsupervisedDenoising2020}, \citet{ buchholzDenoiSegJointDenoising2020} and  \citet{wenDenoisingTrainingTestTime} showed that denoising is also an effective pretext task for segmentation.  \citet{brempongDenoisingPretrainingSemantic2022} however found that combining a frozen pretrained encoder with denoisining pretraining of the decoder was more effective than pretraining the whole model using denoising, except when the number of labeled samples available for fine-tuning is very low. The authors thus recommend to build a segmentation model using an encoder pretrained on Imagenet, keep it frozen, and then pretrain the decoder with denoising. 

Numerous other self-supervised pretraining methods have also been developed for generic encoder pretraining and can be effective for segmentation tasks. Generative pretraining methods are not limited to denoising but also include masked autoencoding \citep{heMaskedAutoencodersAre2021}. Discriminative methods such as SimCLR \citep{chenSimpleFrameworkContrastive2020}, MoCo \citep{heMomentumContrastUnsupervised2020} or Dino \citep{caronEmergingPropertiesSelfSupervised2021} have also proven to be very effective but require to design manually the set of transformations that is used to build the different views of an image.  These pretraining methods generally only allow to get low resolution feature maps, which may be suboptimal for segmentation tasks, but some pretext tasks for self-supervised learning have been specifically developed for dense prediction models such as  pixel-wise contrastive learning \citep{wangDenseContrastiveLearning2021a}.

\citet{NIPS2001_7b7a53e2} provided in 2001 theoretical justifications that training a model to learn the joint distribution $p(x,y)$ allows for label-efficient training. \citet{NIPS2003_b53477c2} introduced the expression "hybrid model" to designate models  trained using both generative and discriminative targets. In the context of image classification,
\citet{khoslaSupervisedContrastiveLearning2021}, \citet{suAllinOnePreTrainingMaximizing2023b} and Liang et al \citet{liangSupMAESupervisedMasked2024} have proposed image encoder pretraining methods involving both self-supervised learning and supervised learning from image class labels. Closer to our work, \citet{dejaLearningDataRepresentations2023b} proposed to pretrain a shared encoder jointly for diffusion  pretraining and supervised noisy image classification. 

Several works have built joint generative models associated to segmentation datasets and studied their possible applications. Most of these works \citep{toker2024satsynth,9577984,DBLP:conf/cvpr/LiLKKF022} proposed to use this joint generative model to produce new training samples, but 
\citet{liSemanticSegmentationGenerative2021} built a GAN generative model for the joint distribution of images and masks, and showed that using the associated latent space could be effective for generalization and label-efficient learning. 

Numerous works \citep{10.1007/978-3-030-01219-9_18, Chen2019DomainAF,Liu2021SourceFreeDA} have addressed the problem of adapting a segmentation model trained on one domain to another related domain. Except for fine-tuning the pretrained model on the available labeled samples, the methods developed for this task however generally assume that a large number of unlabeled samples is available in the target dataset.

\section{Methodology}
\subsection {Hybrid diffusion models}

Let us first recall why an image denoiser conditioned on the noise level can be considered as a generative model thanks to the theory of diffusion models.
We follow the theorical framework of \citet{songScoreBasedGenerativeModeling2021}, the notations of \citet{kingmaVariationalDiffusionModels},
and consider a discrete forward noising process $x_0, x_1,...,x_T$ which, if T is large, can be identified with a stochastic process described by the stochastic differential equation
\begin{equation}
  dx_t = f(t)x_tdt + g(t)dw \;,
\end{equation}

 with 
  $q_t(x_t | x_0) = \mathcal N(x_t | \alpha_t x_0, \sigma_t^2 I) $ and $f(t) = \frac {d \log \alpha_t}{dt} $, $g^2(t) = \frac {d \sigma_t^2}{dt} - 2 \frac {d \log \alpha_t}{dt} \sigma_t^2$.

The associated reverse SDE \citep{andersonReversetimeDiffusionEquation1982} is
\begin{equation}
dx_t = [f(t)x_t -g(t)^2 \nabla_{x_t} \log(p(x_t))]dt + g(t) d \bar w  \;,
\end{equation}
where $ d \bar w$ refers to  a reverse brownian motion.

The quantity $\nabla_{x_t} \log(p(x_t))$ can be learnt from the data thanks to Tweedie's formula \citep{efronTweedieFormulaSelection2011}, which leads to the equality
\begin{equation}
  \nabla_{x_t} \log(p(x_t))  = \frac {  \alpha_t \mathbb E [ x_0 | x_t] -x_t } {\sigma_t^2}\;. 
  \end{equation}
As a consequence, the reverse SDE can be written as 
\begin{equation}\label{eq:rev_sde}
 dx_t = [f(t)x_t -g(t)^2 (\frac {  \alpha_t \mathbb E [ x_0 | x_t] -x_t } {\sigma_t^2} )]dt + g(t) d \bar w\;, 
 \end{equation}
and the associated probability flow \citep{songScoreBasedGenerativeModeling2021} can be described by the ODE
\begin{equation} \label {eq:rev_ode} dx_t = [f(t)x_t -\frac {g(t)^2}2 (\frac {  \alpha_t \mathbb E [ x_0 | x_t] -x_t } {\sigma_t^2} )]dt\;.  
\end{equation}

Training a denoiser conditioned on the noise level  or on $t$ using an MSE loss provides an estimator of $\mathbb E [ x_0 | x_t]$ and allows one to get a complete description of the reverse SDE or ODE. Assuming that $\alpha_T$ is small compared to $\sigma_T$ so that $p(x_T)$ can be identified with pure Gaussian noise, it is then possible to generate samples from the distribution $p(x)$ by  sampling $x_T$ from a Gaussian distribution and then solving the equation \ref{eq:rev_sde}  or \ref{eq:rev_ode} using an SDE or ODE solver.

Let us now consider a dataset composed of pairs $z_i = (x_i,y_i)$, where the $x_i$ are i.i.d. following some unknown distribution $p(x)$, and  $y$ is a deterministic unknown function of $x$, $y = \mu(x)$. In this paper, $x$ will be an image that we want to analyse and $y$ the associated segmentation mask.

We propose to train a  denoiser conditioned on $t$ to provide an estimate of $x$ and $y$ from a noisy version $x_t$ of $x$ only. 
 If we assume that we train this denoiser with an MSE loss, the denoiser provides an estimator of $\mathbb E[z | x_t] = \mathbb E [x,y | x_t]$. 
 
 This denoiser can be trivially used to generate samples from the distribution $p(x,y)$ by first using the $x$ estimates $\mathbb E [x| x_t]$ of the denoiser  to generate samples from the  distribution $p(x)$ using the associated ODE or SDE, and then using the denoiser at step zero to predict $y$ from $x$, since for $t=0$ (no noise), the trained denoiser provides an estimate of $\mathbb E [y| x] = \mu(x)$

It appears however than we can also generate samples from the joint distribution $p(x,y)$ by solving a reverse SDE or ODE involving this denoiser, which is formally fully similar to the reverse SDE or ODE of diffusion models described above. More precisely, we have the following proposition for the SDE: 

\paragraph{Proposition 1} \itshape Let us assume that $x_T$ is sampled from a normal Gaussian distribution and $y_T$ is set to any arbitrary value, and let $z_t = (x_t, y_t)$. Then, solving the reverse SDE
\begin{equation}
\label{eq:sde_hybrid}
 dz_t = [f(t)z_t -g(t)^2 (\frac {  \alpha_t \mathbb E [ z | x_t] -z_t } {\sigma_t^2} )]dt + g(t) d \bar w 
 \end{equation}allows one to obtain samples from the distribution $p(z) = p(x,y)$ for $t=0$. \upshape

It can be noted that equation \ref{eq:sde_hybrid}  is exactly the same  as equation \ref{eq:rev_sde} except that the expectation term is $\mathbb E [ z | x_t]$ and not $\mathbb E [ z | z_t]$.
We have a similar result for the probability flow ODE: 

\paragraph{Proposition 2}
\itshape Let us assume that $x_T$ is sampled from a normal Gaussian distribution and $y_T$ is set to any arbitrary value, and let $z_t = (x_t, y_t)$. Then solving the ODE
\[ dz_t = [f(t)z_t -\frac {g(t)^2}2 (\frac {  \alpha_t \mathbb E [ z | x_t] -z_t } {\sigma_t^2} )]dt \]
allows to obtain samples from the distribution $p(x,y)$ for $t=0$.\upshape

The proofs of these propositions are provided in the supplementary material.

One may intuitively apprehend Proposition 2 by considering that the ODE associated to standard diffusion models provides a mapping between data samples $(x,y)$ and latent variables $(x_T,y_T)$ following a Gaussian distribution. If $y$ is a deterministic function of $x$, this leads to an unnecessary increase of the dimensionality of the latent space compared to the dimensionality of the data manifold. Proposition 2 shows that it is possible to avoid this augmentation by reducing the dimensionality of the latent space to the dimensionality of the image space.

We then call a noise conditional model providing an estimate of $\mathbb E [ z | x_t] = \mathbb E[ x,y | x_t] $ a hybrid diffusion model: It can be considered both as a generative model for the joint distribution $p(z) = p(x,y)$, but also, taking $t = 0$, as a pure discriminative segmentation model.

\subsection{Using hybrid diffusion models for representation learning}

Considering that noise-conditioned denoisers used for diffusion models have shown to be effective representation learners in the unsupervised setting, we expect a hybrid diffusion model to also be  effective for representation learning. Since the hybrid diffusion model has learned the full joint distribution of the segmentation dataset,  we expect the representations learned by a hybrid diffusion model  to be more effective than those learned using pure unsupervised learning or pure supervised learning.

To perform label-efficient transfer learning, we then propose to first train a hybrid diffusion model on a first domain where we have a lot of labeled samples, and then fine-tune it on a second domain, where only a few samples are available. The architecture of a hybrid diffusion model is  the same as  a standard unconditional diffusion model, with the exception that the number of output channels has to be extended for color images from 3 to $3+K$, where $K$ is the number of segmentation classes since we use one-hot encoding for the segmentation masks.

We use the same standard MSE loss function  as in \citep{salimansProgressiveDistillationFast2022} except that we have found from our experiments that it is necessary to insert a weighting factor $\lambda$ to give a significantly lower weight to the segmentation masks reconstruction loss. If $x,y$ are the ground-truth images and masks and $\hat x, \hat y$ the predicted images and masks, the loss function is then

$$ \mathcal L(x,y, \hat x, \hat y,t) = \frac {\alpha_t^2}{\sigma_t^2}( \frac 1{3hw}\norm{x - \hat x}^2 + \frac 1{Khw}\lambda \norm{y -\hat y }^2)\;, $$
where $h,w$ are  the height and width of the image and $K$ is the number of segmentation classes.

We will investigate the two fine-tuning techniques that have been introduced in \ref{sec:related_works} and have proven to be effective for diffusion models: 
\begin{itemize}
\item The first one, which we will call vanilla fine-tuning, is to freeze the condition $t=1$ and perform a supervised fine-tuning of this model considered as a segmentation model using the labeled samples of the second domain. This can be considered as a straightforward extension of the PTDR fine-tuning method proposed in \citep{rousseauPreTrainingDiffusionModels2023}. If the number of classes of the second dataset is not the same as the number of classes in the first domain, we replace the last  layer of the UNet by a new layer with the correct number of output channels, initialized randomly.

\item We also adapt the LEDM method \citep{baranchukLabelEfficientSemanticSegmentation2021} to hybrid diffusion models, which is also straightforward since the model we use has the same architecture \citep{dhariwalDiffusionModelsBeat2021}: We freeze the model after pretraining and train a pixel-wise MLP to perform segmentation using as input the feature map activations of the model for different time-steps, rescaled as necessary to the input image size. 
\end{itemize}

\section{Experiments}

\subsection{Datasets} We consider the following segmentation datasets:
\begin{itemize}
\item Skin lesion segmentation: We use the ISIC 2018 dataset  for pretraining and the DermIS  and PH2  datasets for fine-tuning. 
\item Chest X-ray segmentation: We use the Shenzhen dataset  for pretraining and the NIH  and Montgomery  datasets for fine-tuning. 
\item Face segmentation: We use the Celebamask-HQ dataset for pretraining and FFHQ 34 for fine-tuning. It should be noted that while these datasets are closely related (they both show human faces), they do not use the same classes.
\end{itemize}
The main characteristics of these datasets are listed in Table \ref{tab:datasets} and samples of each dataset are provided in the supplementary material.

\begin{table}[ht]
\label{tab:datasets}
\centering
\resizebox{\textwidth}{!}{
\setlength\tabcolsep{4.0pt}
\renewcommand{\arraystretch}{1.4}
\begin{tabular}{lcccc}
      \toprule
      Dataset & used for &  number of samples & number of classes & data augmentation \\
      \midrule
      ISIC 2018 \citep{codellaSkinLesionAnalysis2018}& pretraining &   2594 & 2 & horizontal and vertical flip \\
      DermIS \citep{glaisterAutomaticSegmentationSkin2013}& fine-tuning & 69 & 2 & no \\
      PH2 \citep{mendoncaPH2DermoscopicImage2013}& fine-tuning  & 200 & 2 & no \\
      \midrule
      Shenzhen \citep{jaegerTwoPublicChest,stirenkoChestXRayAnalysis2018}&pretraining&  566 & 2 & random cropping\\
      NIH \citep{tangXLSorRobustAccurate}& fine-tuning &  100 & 2 & no  \\ 
      Montgomery \citep{stirenkoChestXRayAnalysis2018} & fine-tuning & 138 & 2 & no \\
      \midrule
      CelebaMask-HQ \citep{leeMaskGANDiverseInteractive2020}& pretraining & 30000 & 19 &  no \\
      FFHQ-34 \citep{baranchukLabelEfficientSemanticSegmentation2021} & fine-tuning &   40 & 34 & no \\ 
      \bottomrule
\end{tabular}}
\caption{Main characteristics of the datasets used in our evaluation.}

\label{tab:dataset_splits}
\end{table}
We use the full datasets for unsupervised or hybrid pretraining, but only 20 images as training set for fine-tuning, the remaining being used for validation and testing.

For skin lesion datasets, we do not perform any preprocessing except rescaling to $256 \times 256$. For Celebamask-HQ, we observe that the provided masks overlap since the "skin" masks cover the whole face. We then convert the provided masks into non overlapping masks by giving the skin class a lower priority compared to the other classes. For chest X-ray datasets, we perform histogram equalization and gamma correction as in \citet{ohDeepLearningCOVID192020} and convert the one channel images to three channel images by duplicating each channel.

We do not perform any data augmentation on the fine-tuning datasets since our goal is to study the behavior of our model in the low data regime.  On ISIC 2018, we use horizontal and vertical flips. Considering the small size of the Shenzhen dataset and the fact that horizontal or vertical flips would obviously lead to out of distribution samples, we perform limited random cropping: The coordinates of the cropped rectangles are randomly picked at a horizontal or vertical distance to the border of the original image from 0\% to 5\% of the image size  before rescaling to 256$\times$ 256 (Some augmented samples are provided in the supplementary material).

\subsection{Methods}

We use ADM \citep{dhariwalDiffusionModelsBeat2021} \footnote{\url{https://github.com/openai/guided-diffusion}} as UNet architecture for the diffusion model. This choice is motivated by the fact that the LEDM method is designed to use the feature maps of this model.

To limit training costs for this preliminary study, we significantly depart from the settings recommended in \citep{dhariwalDiffusionModelsBeat2021} and also implemented in \citep{baranchukLabelEfficientSemanticSegmentation2021}: We reduce the number of channels of the model from 256 to 128, reducing the number of parameters of the model from 554 M to 138 M parameters. The  number of iterations during pretraining is set to 40'000 with a batch size of 64 for ISIC 2018 and Shenzhen, and 90'000 with a batch size of 64 for Celabamask-HQ, instead of the range 200K-500K with a batch size of 256 recommended in \citep{dhariwalDiffusionModelsBeat2021}.

We pretrain the models using the Hugging Face diffusers framework \citep{von-platen-etal-2022-diffusers}, using the AdamW optimizer with a constant learning rate of $10^{-4}$ and 2000 linear warm-up steps. 

For the fine-tuning phase, we  use the cross-entropy loss, a  batch size of 12 and a learning rate of $2.10^{-4}$. We evaluate the model on the validation set at the end of each epoch using the target metric and stop the training when no improvement is obtained during 20 epochs, keeping the model with the best evaluation score. The results provided are averages over three fine-tuning runs.

For the implementation of the LEDM fine-tuning technique, we use the official code base \footnote{\url{https://github.com/yandex-research/ddpm-segmentation}}. We  perform the tests with  1 MLP  and do not engage in any hyperparameter optimization.

For medical segmentation datasets, we use the average Jaccard index as evaluation metric (average of foreground class IoU per sample).
For the face segmentation datasets, we use the average mIoU per class, following \citep{baranchukLabelEfficientSemanticSegmentation2021}.

$\lambda$ is set to $1.10^{-4}$ following preliminary tests with $ \lambda \in \{1, 10^{-1}, 10^{-2}, 10^{-3}, 10^{-4},10^{-5} \}$.

Other experimental settings and hyperparameters are detailed in the supplementary material.

\subsection{Results and discussion}

Some samples generated using the proposed hybrid diffusion models are provided in the supplementary material and show that the trained hybrid models are genuine generative models.

To assess the performance of the proposed scheme for fine-tuning, we also fine-tune segmentation models on the same fine-tuning datasets according to the following scenarios: 
 \begin{itemize}
 \item training an ADM UNet with random initialization on the fine-tuning datasets (no pretraining at all).  
  \item training a Deeplabv3 model \citep{Chen2017RethinkingAC} with ResNet-50 
backbone pretrained on Imagenet 1K 
 \item training a Segformer\citep{NEURIPS2021_64f1f27b} B3  model with Mit encoder pretrained\footnote{\url{https://huggingface.co/nvidia/mit-b3}}  on Imagenet 1K 
 \item pretraining a diffusion model using the images of the pretraining dataset and fine-tuning with the fine-tuning datasets using the LEDM method
 \item pretraining a diffusion model using the images of the pretraining dataset and fine-tuning with the fine-tuning datasets using vanilla fine-tuning
 \item pretraining a hybrid diffusion model using the methodology described in this paper and fine-tuning it using the LEDM method
 \item pretraining a hybrid diffusion model using the methodology described in this paper and fine-tuning it using vanilla fine-tuning
 \item training a Deeplabv3 model with ResNet-50 backbone pretrained on Imagenet 1K with the pretraining dataset, then fine-tuning it with the fine-tuning dataset
  \item training a Segformer B3 model with Mit encoder pretrained on Imagenet 1K with the pretraining dataset, then fine-tuning it with the fine-tuning dataset
 \end{itemize}
 The results of these evaluations are shown in Table \ref{tab:evaluation_results}.

One can observe that when no specific pretraining dataset is available, the models using encoders pretrained on Imagenet generally give significantly better results than the ADM UNet with random initialization. 

Unsupervised pretraining the ADM UNet as a diffusion model gives a significant boost to the fine-tuning results, confirming the effectiveness of the methods proposed in \citep{baranchukLabelEfficientSemanticSegmentation2021} and \citep{rousseauPreTrainingDiffusionModels2023}. 

Supervised pretraining followed by vanilla fine-tuning also allows to improve the results compared to random initialization, which was also expected, but using the LEDM fine-tuning method with an ADM UNet pretrained using supervised training does not seem to be effective. 

Concerning the results of the hybrid diffusion model proposed in this paper, we observe that the LEDM fine-tuning method is also not effective, except for the FFHQ 34 dataset, which is consistent with the results observed for supervised pretraining.

We observe however that vanilla fine-tuning of the proposed hybrid diffusion model always gives better average results compared to supervised pretraining or unsupervised pretraining, with significant improvements on the DermIS and FFHQ 34 datasets.

Finally, we note that the results of vanilla fine-tuning our proposed hybrid diffusion model are competitive with, or superior to, those of the DeepLabv3 and SegFormer B3 models pretrained on both ImageNet-1K and the pretraining datasets.

\begin{table}[ht]
\label{tab:evaluation_results}
\centering
\resizebox{\textwidth}{!}{
\begin{tabular}{p{5cm}p{5cm}ccccc}
      \toprule
       &pretraining dataset & ISIC 2018 & ISIC 2018		& Shenzhen & Shenzhen  &  CelebaMask-HQ\\
       
       &fine-tuning dataset  &  PH2 20 samples & DermIS 20 samples  & NIH 20 & Montgomery 20 & FFHQ-34 20 samples\\
       &metric			   & Jaccard index & Jaccard index &  Jaccard index & Jaccard index & average of class IoU  \\
    \midrule
    
       no pretraining 
       & ADM UNet &84.53 $\pm 0.68$ &75.74 $\pm 0.41$  & 73.42 $\pm 2.41$ & 91.70 $\pm 2.19$ & 32.95 $\pm 0.94$ \\
      
      \midrule
         Imagenet 1K pretraining 
         &Deeplabv3 model R50  & 82.89 $\pm 0.83$  & 73.89 $\pm 4.16$  &90.54 $\pm 0.42$ & 93.97 $\pm 0.13$ & 40.00 $\pm 1.16$\\
         
        & Segformer B3 model  & 86.01 $\pm 0.94$ & 86.63 $\pm 1.06$  & 91.19$\pm 0.48$ & 94.90 $\pm 0.29$ & 46.52 $\pm 0.33$\\
        
    \midrule

    unsupervised pretraining 
    &LEDM on diffusion model & 86.07 & 81.83 & 87.03&  90.94 & 54.44\\
    
     &fine-tuning of diffusion model  & 85.08 $\pm 0.63$ & 79.67 $\pm 2.43$ &  87.65 $\pm 0.45$ & 94.31 $\pm 0.44$ &  55.08 $\pm 0.25$\\

      \midrule

      pretraining on labeled dataset 
      &LEDM on supervised ADM Unet 
      & 84.09 & 78.01  &89.33  & 91.46 &45.66\\
       &fine-tuning of supervised ADM UNet &87.94 $\pm 0.42$  & 81.45 $\pm 0.16$  & 90.68 $\pm 1.86$ & 94.54 $\pm 1.23$ & 54.49 $\pm 0.63$ \\
       &LEDM on hybrid model & 85.43 & 78.16  & 85.92 & 90.71& 56.59 \\
     & fine-tuning of hybrid model & 88.23 $\pm 0.21$ & 86.76 $\pm 1.04$  & 92.17 $\pm 0.45$ & 95.45 $\pm 0.11$ & 56.91 $\pm 0.56$ \\
      
            \midrule
       I1K pretraining + supervised pretraining 
         & fine-tuning of supervised Segformer model & 87.74 $\pm 0.37$ & 88.28 $\pm 0.17$ & 92.36 $\pm 0.10$ & 95.66 $\pm 0.19$ & 51.06 $\pm 0.27$\\
       
    & fine-tuning of supervised Deeplabv3 model & 89.44 $\pm 0.36$ & 86.50 $\pm 0.46$  & 92.22 $\pm 0.14$ & 95.55 $\pm 0.05$ & 47.32 $\pm 0.37$\\
   
      \bottomrule
\end{tabular} 
}
$\quad$
\caption{Empirical results.}
\end{table}

\section{Conclusion}

We have introduced in this paper the notion of hybrid diffusion model and provided experimental evidences that vanilla fine-tuning of such a model can be more effective than fine-tuning a similar supervised model or pretrained diffusion model. Further work is needed to explore the full potential of this new paradigm, but  the preliminary experimental results described in this paper show that it allows one to benefit simultaneously from both supervised and unsupervised pretraining for segmentation tasks, which seems to be a natural requirement in many situations where labeled data is available in one domain and we are interested to adapt a model to a new domain where only a few samples are available.


\bibliographystyle{plainnat}

\bibliography{biblio}


\appendix

\section{Supplementary material}

\subsection{Proofs}

Since propositions 1 and 2 are already known for classical diffusion models, we only prove the propositions for $y_t$ and assume that the reverse ODE or SDE applied to $x_t$ allows to sample $x_0$ values from the distribution $p(x)$.

We recall that $$f(t) = \frac {d \log \alpha_t}{dt} $$ and $$g^2(t) = \frac {d \sigma_t^2}{dt} - 2 \frac {d \log \alpha_t}{dt} \sigma_t^2 = 2\sigma_t^2 \frac {d}{dt} \log( \frac {\sigma_t}{\alpha_t}) $$

\subsubsection {Proof of proposition 1 (reverse SDE)}

We consider the reverse SDE for $y_t$ given by the equation
$$
dy_t = [f(t)y_t +\frac {g(t)^2} {\sigma_t^2} ( y_t - \alpha_t \mathbb E[y | x_t] )]dt + g(t) d \bar w \;. 
$$

We recognize a an Ornstein-Ulhenbeck process with time-dependent parameters, which can be solved using the method of variation of constants.
Considering that
$$
f(t) + \frac {g(t)^2} {\sigma_t^2} = \frac d{dt} \log ( \frac {\sigma_t^2}{\alpha_t}) \;, 
$$

the SDE can the be written as 
$$ dy_t = \frac d{dt} \log ( \frac {\sigma_t^2}{\alpha_t})  y_t dt  - 2\frac d{dt} (\log( \frac {\sigma_t}{\alpha_t})) \alpha_t \mathbb E[y | x_t]dt + g(t) d \bar w \;.  
$$
Since $\frac {\sigma_t^2}{\alpha_t}$ is a solution to the associated homogeneous equation, we apply Ito's formula to the function $\phi(t,y_t) = y_t \frac {\alpha_t}{\sigma_t^2}$ and get  

$$ y_T \frac {\alpha_T}{\sigma_T^2} -  y_\tau \frac {\alpha_\tau}{\sigma_\tau^2} =   \int_\tau^T \frac {\alpha_t}{\sigma_t^2} dy_t  + \int_\tau^T \frac d{dt} ( \frac {\alpha_t} {\sigma_t^2}) y_t dt $$

$$ = \int_\tau^T \frac {\alpha_t}{\sigma_t^2} (\frac d{dt} \log ( \frac {\sigma_t^2}{\alpha_t})  y_t dt  - 2\frac d{dt} (\log( \frac {\sigma_t}{\alpha_t})) \alpha_t \mathbb E[y | x_t]dt + g(t) d\bar w  ) + \int_t^T \frac d{dt} ( \frac {\alpha_t} {\sigma_t^2}) y_t dt $$

$$  = \int_\tau^T \frac d{dt} ( \frac {\alpha_t^2}{\sigma_t^2}) \mathbb E[y | x_t]dt + \int_\tau^T \frac {\alpha_t}{\sigma_t^2} g(t)d \bar w \;. $$


Considering that $\alpha_T = 0$, we get

$$ y_\tau = - \frac {\sigma_\tau^2}{\alpha_\tau}\int_\tau^T \frac d{dt} ( \frac {\alpha_t^2}{\sigma_t^2}) \mathbb E[y | x_t]dt  -\frac {\sigma_\tau^2}{\alpha_\tau} \int_\tau^T \frac {\alpha_t}{\sigma_t^2} g(t)d \bar w \; .$$

 We remark that the stochastic term, which we note $S_\tau$, converges almost surely to zero as $\tau$ converges to zero considering that the associated quadratic variation   

$$ \mathbb E[ S_\tau^2] =  \int_\tau^T \frac {\sigma_\tau^4}{\alpha_\tau^2} \frac {\alpha_t^2}{\sigma_t^4} g(t)^2dt = 
  \int_\tau^T \frac {\sigma_\tau^4}{\sigma_t^4} \frac {\alpha_t^2}{\alpha_\tau^2} (\frac {d \sigma_t^2}{dt} - 2 \frac {d\alpha_t}{dt} \sigma_t^2) dt  $$

converges to zero since $\sigma_0 = 0$ and $\alpha_0= 1$.

Concerning the first term, we observe that it only depends on the behavior of $\mathbb E [y | x_t]$ near $t= 0$. Writing $\mathbb E[y | x_t] = \mathbb E[y | x_0] + \epsilon(t)$, this expression becomes

$$ - \frac {\sigma_\tau^2}{\alpha_\tau}\int_\tau^T \frac d{dt} ( \frac {\alpha_t^2}{\sigma_t^2}) \mathbb E[y | x_0]dt- \frac {\sigma_\tau^2}{\alpha_\tau}\int_\tau^T \frac d{dt} ( \frac {\alpha_t^2}{\sigma_t^2}) \mathbb \epsilon(t) dt \; .$$

The first term converges to $\mathbb E[y | x_0] = \mu(x_0)$ and the  second term converges to zero  since 

$$ \int_\tau^T \abs{ \frac d{dt}(\frac {\alpha_t^2}{\sigma_t^2})} dt = \frac {\alpha_\tau^2}{\sigma_\tau^2} - \frac {\alpha_T^2}{\sigma_T^2} \;, $$

and $\epsilon(t)$ converges to zero almost surely.

\subsubsection {Proof of proposition 2 (ODE)}

We consider the ODE for $y_t$
$$dy_t = [f(t)y_t +\frac {g(t)^2} {2 \sigma_t^2} ( y_t - \alpha_t \mathbb E[y | x_t] )]dt \;, $$

and have to show that  starting from any value of  $y_T$, this  ODE leads to  the value $y_0 = \mathbb E[y | x_0] = \mu (x_0) $. This ODE is a first order linear equation which can also be solved in integral form using the method of variation of constants:

We have

$$ f(t) + \frac {g(t)^2} {2 \sigma_t^2} = \frac {\dot \sigma_t}{\sigma_t} \; ,$$

and the ODE can then be written as:
$$ \dot y_t = \frac {\dot \sigma_t}{\sigma_t}y - \frac d{dt} (\log( \frac {\sigma_t}{\alpha_t})) \alpha_t \mathbb E[y | x_t] \;. $$

Considering that $\sigma_t$ is a solution of the associated homogeneous equation, we get

$$ \frac {d}{dt} ( \frac {y_t}{\sigma_t}) =  - \frac {\alpha_t}{\sigma_t}\frac d{dt} (\log( \frac {\sigma_t}{\alpha_t}))  \mathbb E[y | x_t] = \frac d{dt}( \frac {\alpha_t}{\sigma_t}) ) \mathbb E[y | x_t] \;, $$

and 
$$ \frac {y_T}{\sigma_T} =  \frac {y_\tau}{\sigma_\tau} + \int_\tau^T ( \frac d{dt}( \frac {\alpha_t}{\sigma_t}) ) \mathbb E[y | x_t] dt \;, $$
which leads to
$$ y_\tau = \frac {\sigma_\tau}{\sigma_T} y_T - \sigma_\tau \int_\tau^T ( \frac d{dt}( \frac {\alpha_t}{\sigma_t}) ) \mathbb E[y | x_t] dt\;. 
$$
And we conclude by the same argument as for the SDE.

\subsection{Dataset samples and generated samples}

We provide below some samples of the datasets used in this paper as well as some generated samples for the datasets used for pretraining. 
The generated samples are created using trained hybrid diffusion models and the DPM-Solver ODE solver \citep{NEURIPS2022_260a14ac} implemented in the Hugging Face Diffusers framework with 100 inference steps. $x_T$ values are sampled from a Gaussian normal distrution and all the $y_T$ mask values are set to zero.
The generated images shown are the direct outputs of the ODE solver. The generated masks shown are computed by selecting for each pixel the class with the higher mask value.
\newpage
\subsubsection {ISIC 2018 dataset}

\begin{itemize}
\item Dataset samples

\includegraphics[width=\textwidth]{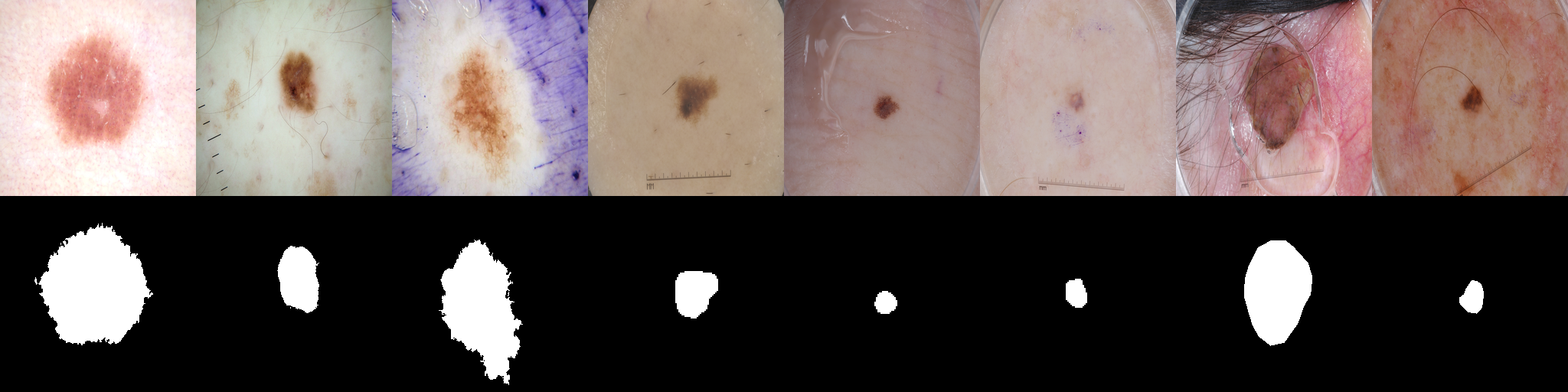}

\item Generated samples

\includegraphics[width=\textwidth]{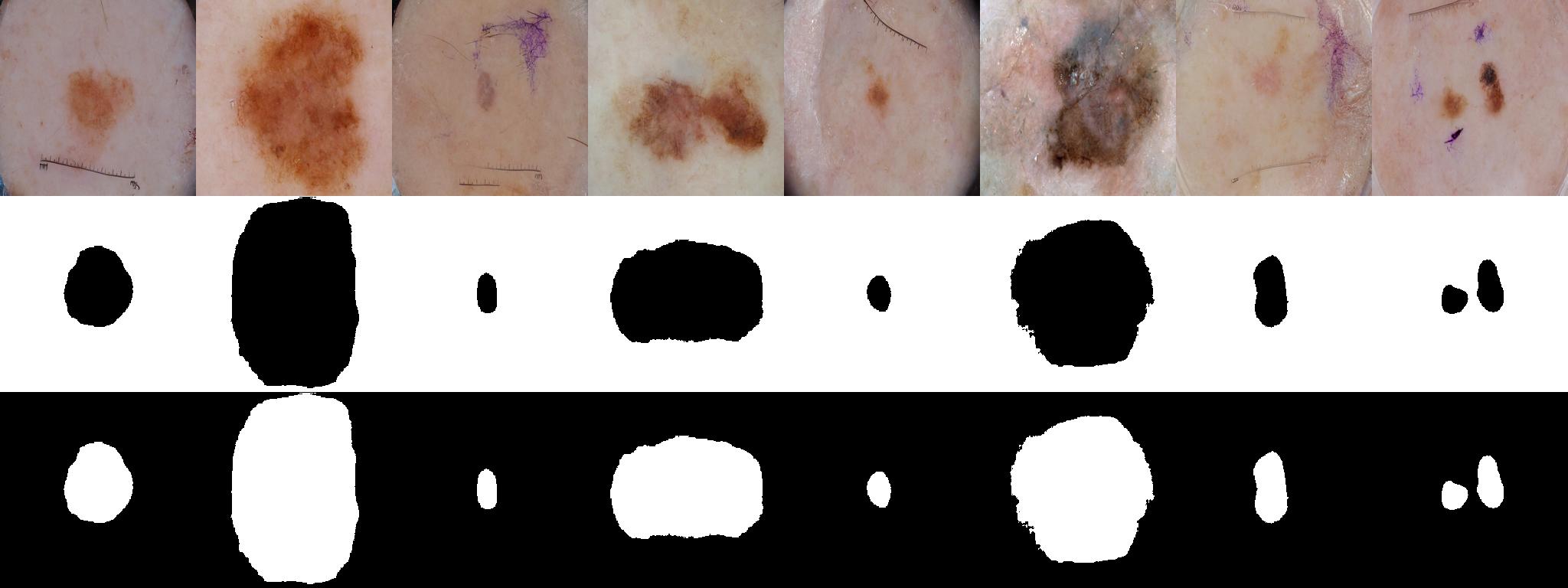}

\end{itemize}
\subsubsection {DermIS  dataset}
\begin{itemize}
\item Dataset samples

\includegraphics[width=\textwidth]{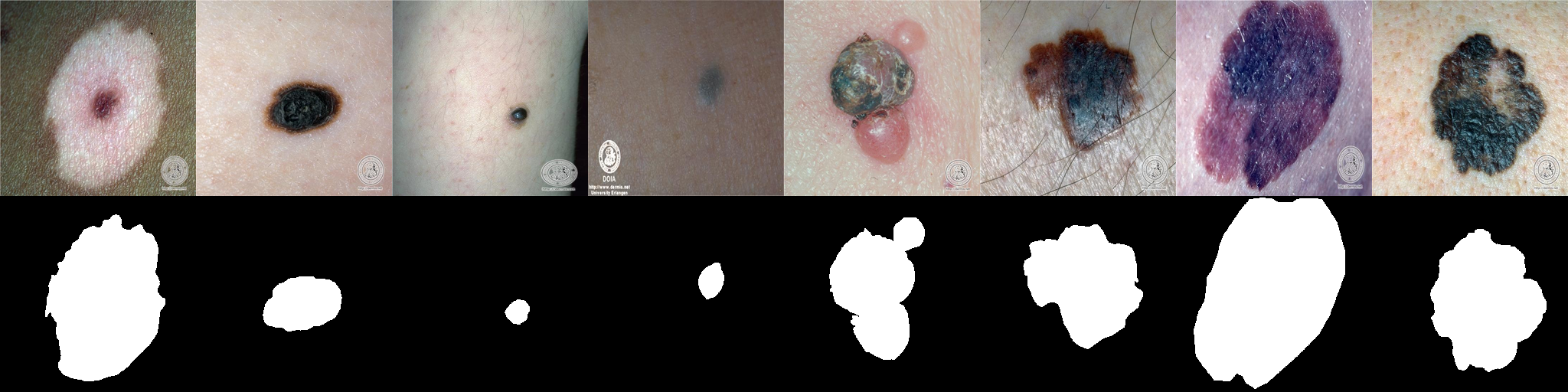}
\end{itemize}

\subsubsection{PH2 dataset}
\begin{itemize}
\item Dataset samples:

\includegraphics[width=\textwidth]{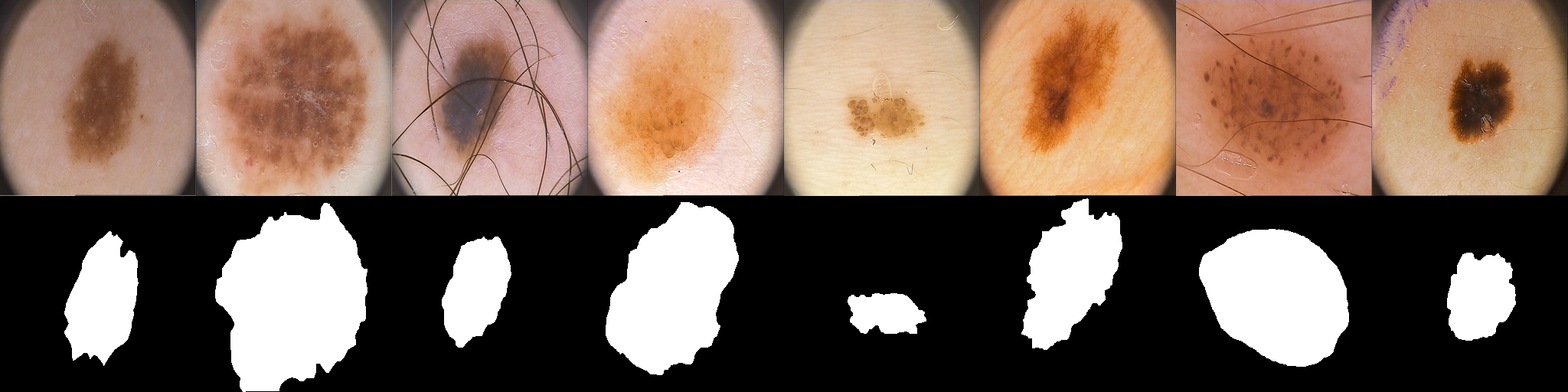}
\end{itemize}

\newpage
\subsubsection {Shenzhen dataset}
\begin{itemize}
\item Dataset samples (no data augmentation):

\includegraphics[width=\textwidth]{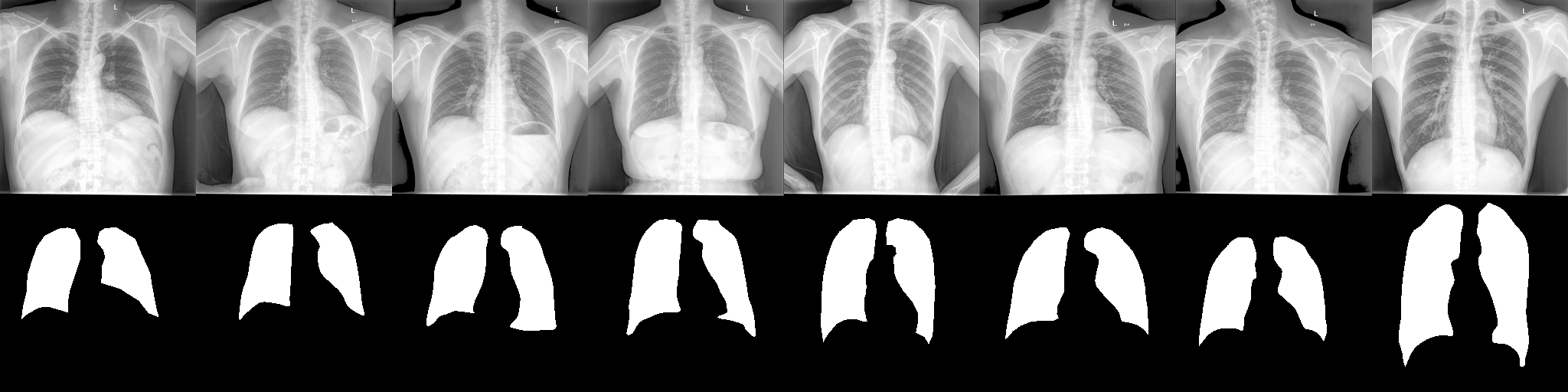}

\item Dataset samples (with data augmentation):

\includegraphics[width=\textwidth]{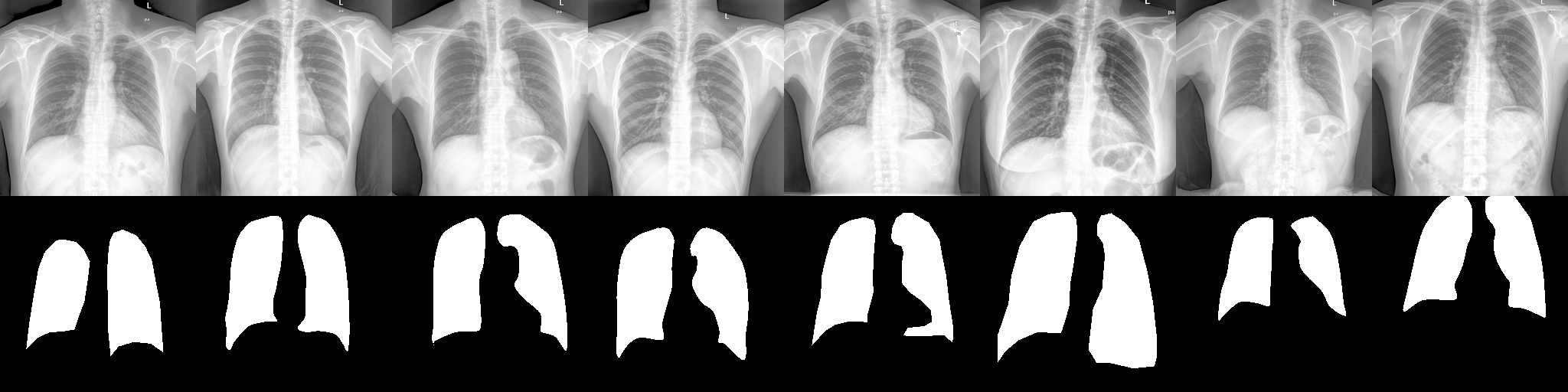}

\item Generated samples

\includegraphics[width=\textwidth]{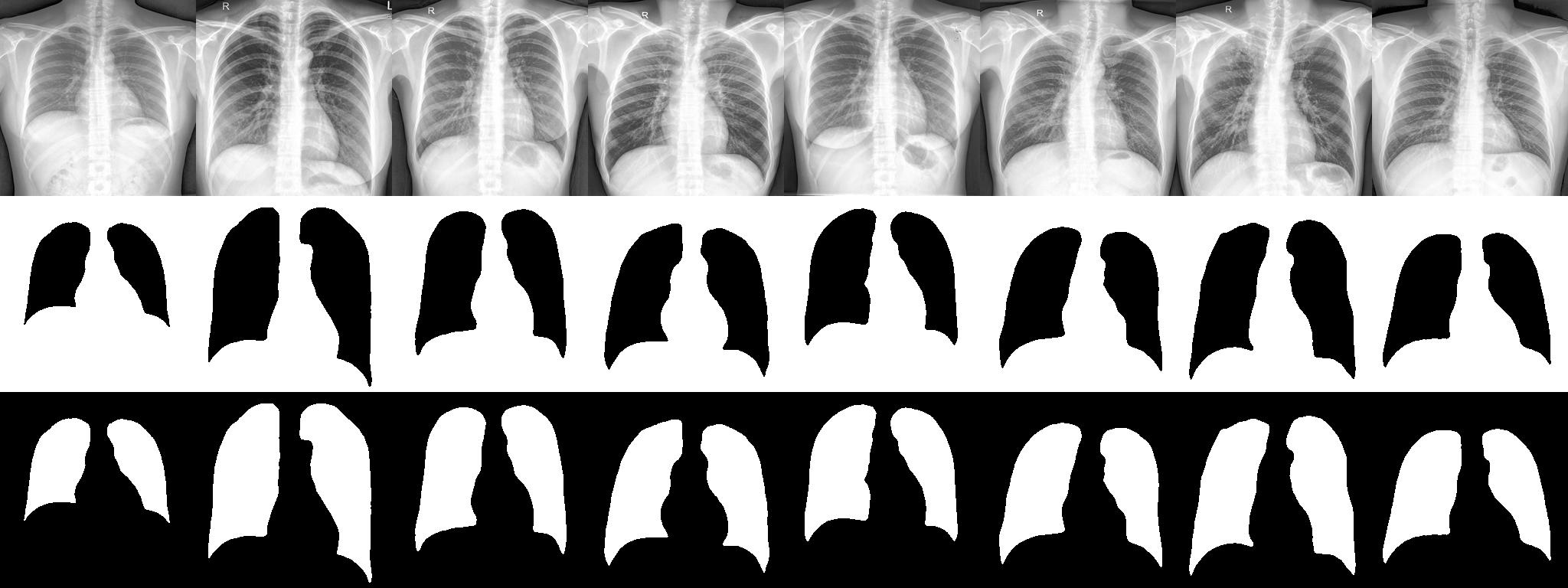}
\end{itemize}
\subsubsection{NIH dataset}

\begin{itemize}
\item Dataset samples:

\includegraphics[width=\textwidth]{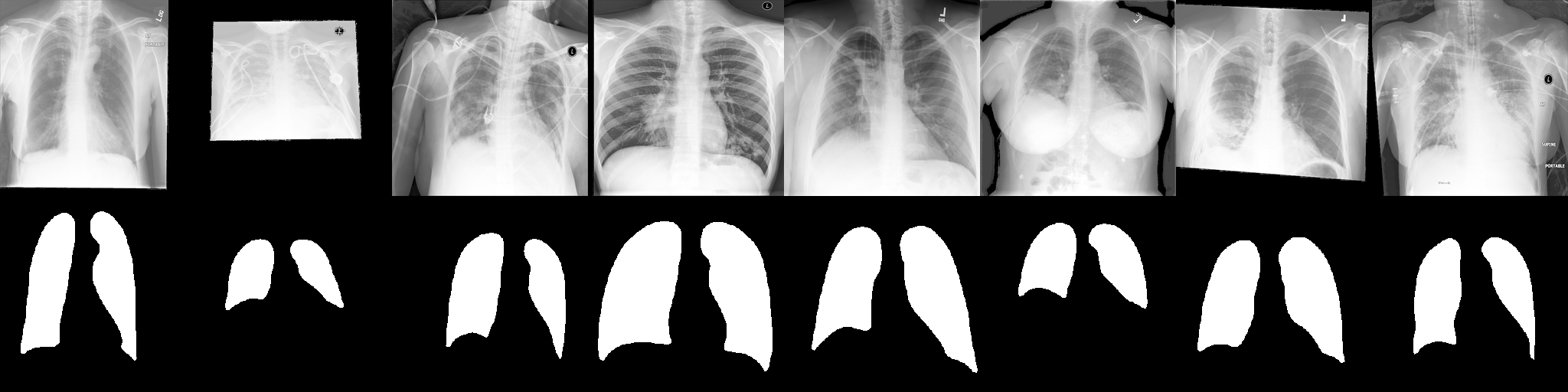}
\end{itemize}
\newpage
\subsubsection{Montgomery dataset}

\begin{itemize}
\item Dataset samples:

\includegraphics[width=\textwidth]{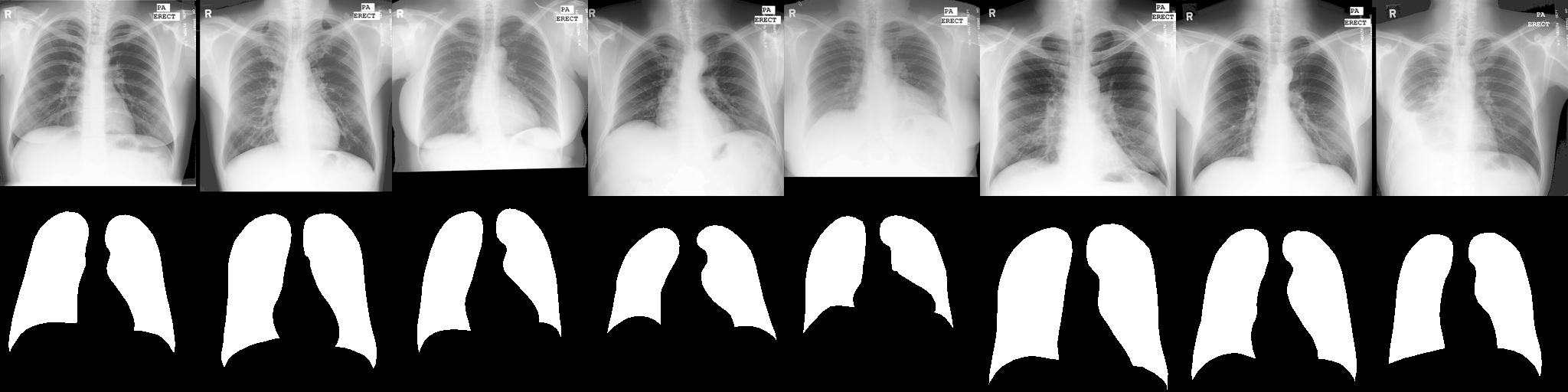}
\end{itemize}

\newpage
\subsubsection {Celebamask-HQ dataset}
\begin{itemize}
\item Dataset samples:

\includegraphics[height=0.95\textheight]{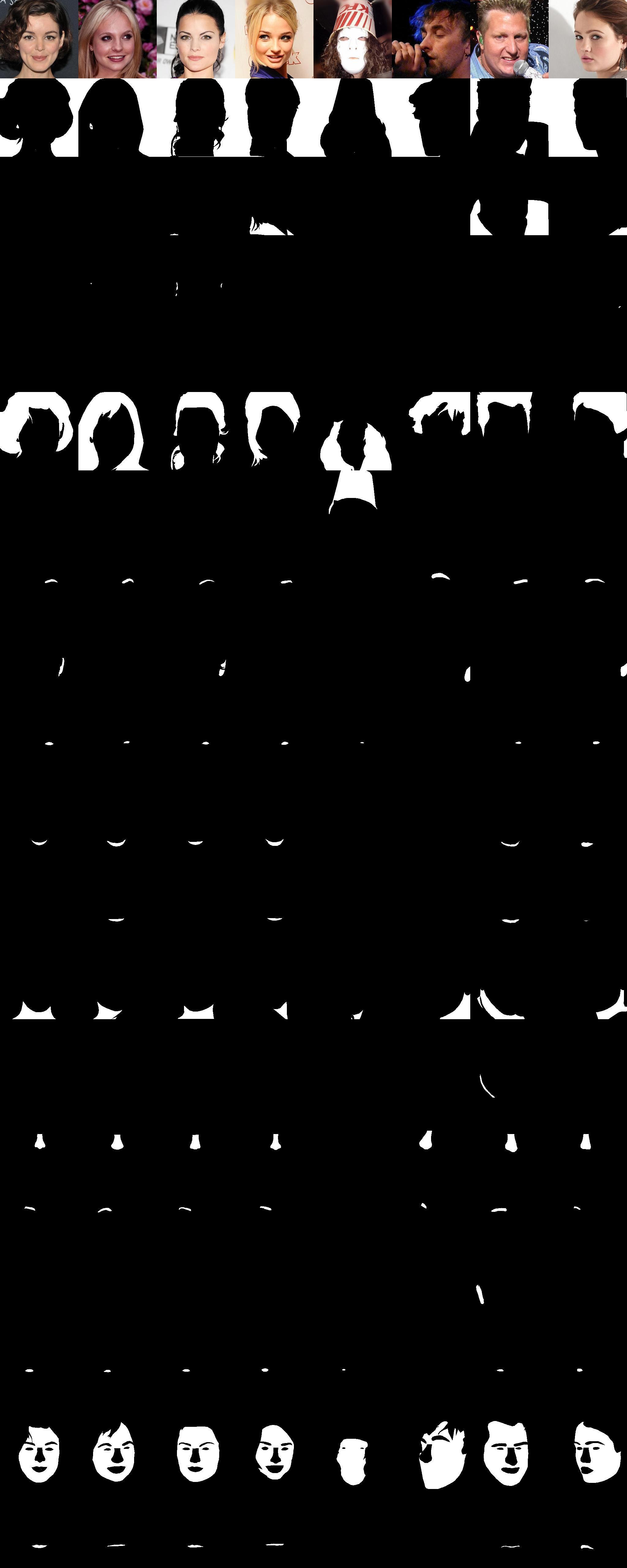}

\newpage
\item Generated samples:

\includegraphics[height=0.95\textheight]{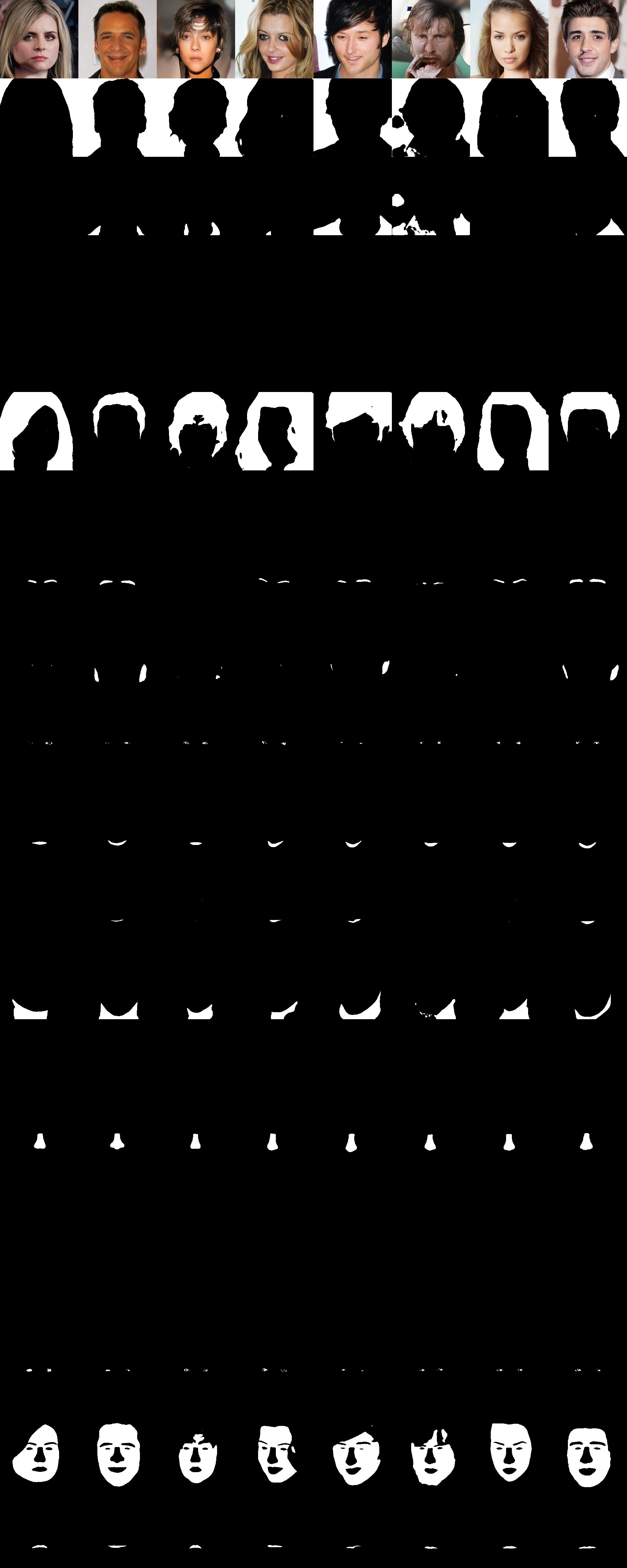}
\end{itemize}
\newpage
\subsubsection {FFHQ-34 dataset}
\begin{itemize}

\item Dataset samples:

\includegraphics[width=\textwidth]{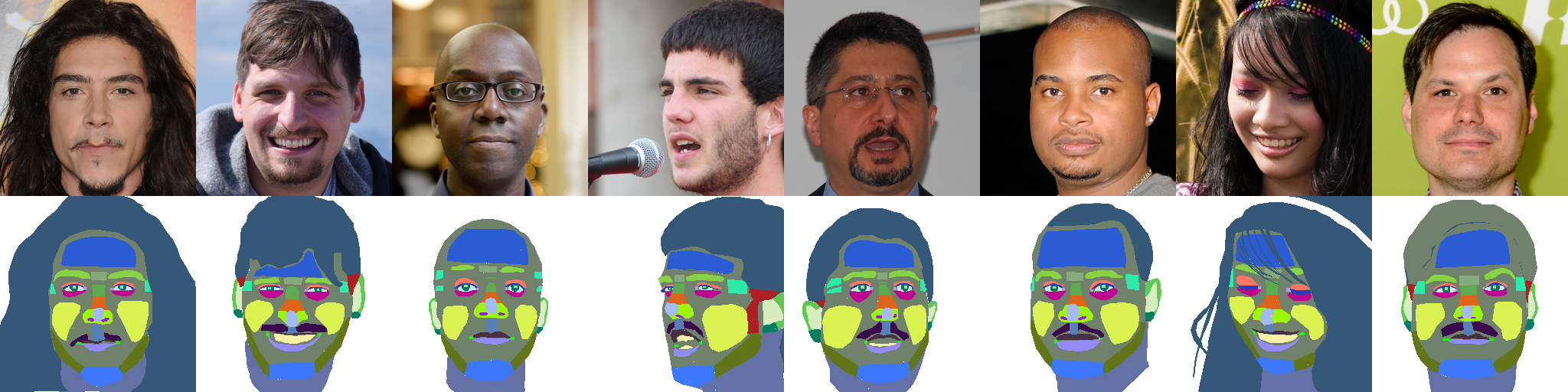}

\end{itemize}

\subsection{Training and architecture details}

The architecture of the ADM Unet is the same as in \citet{dhariwalDiffusionModelsBeat2021}, except for the number of model channels set to 128 instead of 256 and the number of ouput channels set at $3 + K$ where $K$ is the number of classes. Chest X-ray black and white images ( 1 input channel) are converted to 3 channels images by duplicating the input channels. Pixel values are normalized to the range $[-1,1]$ before being given to the Unet. 

The full configuration settings of this ADM Unet in reproduced below for completeness:

\begin{itemize}
\item attention resolutions: [32,16,8]
\item channel multipliers:[1,1,2,2,4,4]
\item dropout : 0.1
\item image size : 256
\item number of input channels:3
\item number of ouput channels:3+K 
\item number of model channels: 128
\item number of head channels: 64
\item number of resisdual blocks: 2
\item resblock updown: true
\item use fp16: false,
\item use scale shift norm: true
\end{itemize}

We keep most of the default hyperparameters of the diffusers framework for training an unconditional diffusion model. We reproduce them below for completeness. It should be noted that since the time indexes associated to nonzero noises used in the diffusers framework are from 0 to 999 and not 1 to 1000, the fine-tuning is done with $t=0$ and not $t=1$.

\begin{itemize}
\item batch size: 64 for pretraining, 12 for fine tuning
\item learning rate: 1e-4 for pretraining, 2e-4 for fine-tuning
\item optimizer: AdamW
\item Adam beta1: 0.95
\item Adam beta2: 0.99
\item Adam weight decay: 1e-6
\item Adam epsilon: 1e-8
\item number of warmup steps: 2000 for pretraining, 0 for fine-tuning
\item Number of DDPM steps (T): 1000
\item Schedule beta type: linear
\item DDPM beta end: 0.02
\end{itemize}

\end{document}